\documentclass[preprint,journal]{vgtc}            

\onlineid{1830}

\vgtccategory{Research}

\title{FA-INR: Adaptive Implicit Neural Representations for Interpretable Exploration of Simulation Ensembles}

\author{%
  Ziwei Li, Yuhan Duan, Tianyu Xiong, Yi-Tang Chen, Wei-Lun Chao, and Han-Wei Shen \\
}

\authorfooter{
  \item
  	Ziwei Li, Yuhan Duan, Tianyu Xiong, Yi-Tang Chen, Wei-Lun Chao, and Han-Wei Shen with The Ohio State University.
  	E-mail: {\{li.5326, duan.418, xiong.336, chen.10522, chao.209, shen.94\}@osu.edu}
}

\usepackage{amsmath,amsfonts,bm}

\def\eqref#1{equation~\ref{#1}}

\def\1{\bm{1}}

\DeclareMathAlphabet{\mathsfit}{\encodingdefault}{\sfdefault}{m}{sl}
\SetMathAlphabet{\mathsfit}{bold}{\encodingdefault}{\sfdefault}{bx}{n}

\def\sD{{\mathbb{D}}}

\usepackage{amssymb}
\usepackage{amsmath,amsfonts}
\usepackage{amsopn}
\usepackage{bm} 
\usepackage{multirow}
\usepackage{flushend}
\usepackage{tabularx}
\usepackage{xspace}

\def\eqref#1{equation~\ref{#1}}

\def\1{\bm{1}}

\DeclareMathAlphabet{\mathsfit}{\encodingdefault}{\sfdefault}{m}{sl}
\SetMathAlphabet{\mathsfit}{bold}{\encodingdefault}{\sfdefault}{bx}{n}

\newcommand{\eat}[1]{}
\newcommand{\method}[1]{\textsc{#1}}

\newcommand{\ours}{\method{FA-INR}\xspace}

\newcommand{\eg}{{\em e.g.}}
\newcommand{\ie}{{\em i.e.}}

\abstract{
Surrogate models are essential for efficient exploration of large-scale ensemble simulations. Implicit neural representations (INRs) provide a compact and continuous framework for modeling spatially structured data, but they often struggle with learning complex localized structures within the scientific fields.
Recent INR-based surrogates address this by augmenting INRs with explicit feature structures, but at the cost of flexibility and substantial memory overhead.
In this paper, we present Feature-Adaptive INR (\ours), an adaptive INR-based surrogate model for high-fidelity and interpretable exploration of ensemble simulations. Instead of relying on structured feature representations, \ours leverages cross-attention over a learnable key-value memory bank to allocate model capacity adaptively based on the data characteristics.
To further improve scalability, we introduce a coordinate-guided mixture of experts (MoE) framework that enhances both efficiency and specialization of feature representations. More importantly, the learned experts produce an interpretable partition over the simulation domain, enabling scientists to identify complex structures and perform localized parameter-space exploration.
Beyond quantitative and qualitative evaluations, we also demonstrate that our learned expert specialization can reveal meaningful scientific insights and support localized sensitivity analysis.
}


\graphicspath{{figs/}{figures/}{pictures/}{images/}{./}} 

\usepackage{tabu}                      
\usepackage{booktabs}                  
\usepackage{lipsum}                    
\usepackage{mwe}                       
\usepackage{ccicons}                   
\usepackage{xcolor}
\usepackage{wrapfig}
\usepackage{makecell}

\usepackage{mathptmx}                  

\begin{document}



\maketitle

\section{Introduction}
\label{sec:intro}


Accurately simulating complex physical systems is essential for scientific discovery, but running high-fidelity simulations at scale is often prohibitively expensive~\cite{chartier2021carpool, schneider2024opinion, wang2025surrogate}. To address this challenge, {Surrogate models} have emerged as a practical alternative.
By learning from simulation outputs or observational data, surrogate models can provide fast and accurate predictions~\cite{he2019insitunet, catalani2024neural, luo2024continuous, serrano2023operator, shi2022vdl, shi2022gnn}. This enables researchers to efficiently explore how physical fields evolve under varying simulation conditions in order to understand parameter sensitivity and identify scientific phenomena.
For example, climate scientists can leverage surrogate models to analyze how ocean temperature changes under different wind stress conditions without repeatedly running computationally intensive simulations.
\par
Among the existing surrogate modeling frameworks, implicit neural representations (INRs) are particularly attractive for scientific visualization~\cite{han2022coordnet, han2025moe, wurster2023adaptively, chen2025explorable}. INRs learns a continuous mapping directly from input coordinates (\eg, spatial locations in the ocean) and simulation conditions (\eg, wind stress amplitudes) to target field values (\eg, temperature), enabling compact modeling and resolution-independent querying at arbitrary locations. These properties make them well-suited for spatially structured scientific data and facilitate efficient exploration of simulation ensembles. However, standard INR formulations, typically based on multilayer perceptrons (MLPs)~\cite{mildenhall2021nerf, serrano2023operator, sitzmann2020implicit, catalani2024neural}, often struggle to represent localized, high-frequency structures due to spectral bias~\cite{rahaman2019spectral, fridovich2022spectral}, which limits their fidelity on complex scientific fields.
\par
Recent INR-based approaches address this limitation by augmenting MLPs with learnable embeddings defined over rigid geometric structures, such as grids or planes, to encode complex spatial variations explicitly \cite{cao2023hexplane, chen2025explorable, fridovich2023k, fridovich2022plenoxels, sun2022direct, sun2022improved}. While these methods can improve reconstruction quality, they also introduce two limitations for the scientific data. 
First, rigid geometric structures \textit{allocate representational capacity uniformly} across 3D space, despite the fact that scientific fields often present highly heterogeneous spatial complexity. 
As a result, such data-independent rigid structures \textit{cannot adapt to the underlying data characteristics,} often allocating excessive capacity to smooth regions while under-representing regions with fine-scale structures.
Second, these explicit structures reduce model compactness, which is one of the key strengths of INRs. Specifically, feature grids \textit{introduce significant memory overhead and scale poorly} with increasing dimensionality and spatial resolution, even when low-rank approximations are used \cite{cao2023hexplane, chen2022tensorf, fridovich2023k}. 
These challenges become more significant for large-scale ensemble data, where spatial features may vary substantially across different parameter settings.
\par
In this work, we propose \underline{\textbf{F}}eature-\underline{\textbf{A}}daptive \underline{\textbf{INR}} (\ours), an INR-based surrogate designed for both high-fidelity modeling and interpretable exploration of simulation ensembles.
Specifically, to address the aforementioned challenges, our \textit{first solution} is to replace the rigid feature structures with cross-attention~\cite{vaswani2017attention} over a learnable key-value memory bank, inspired by the memory tokens in language modeling~\cite{lample2019large, wu2022memorizing, zhang2022linearizing}. Given a query coordinate, the model adaptively retrieves features based on the local data characteristics, rather than relying on fixed feature interpolation over the pre-defined structures. In this case, {cross-attention determines \textbf{what} features should be retrieved for each query}, enabling the model to allocate its capacity adaptively based on the data. This adaptive embedding mechanism improves the flexibility while maintaining a compact representation.
\par
Our \textit{second solution }is to introduce a coordinate-guided mixture-of-experts (MoE) architecture~\cite{chen2022towards, du2022glam, fedus2022review, jacobs1991adaptive, riquelme2021scaling, shazeer2017outrageously}. In this design, each expert maintains its own dedicated memory bank. While attention enables adaptive feature retrieval within a memory bank, MoE partitions the spatial domain through a routing mechanism defined by input coordinates. This routing assigns queries to spatially-specialized experts, allowing different experts to focus on different subregions to better capture heterogeneous spatial structures.
In other words, the MoE architecture determines \textbf{where} representational capacity should be allocated. Unlike prior MoE-based INR methods, our framework does not require any pretraining of expert assignments or sub-INR networks. This design not only improves feature utilization and modeling fidelity, but more importantly, it produces an interpretable spatial decomposition over the simulation domain. Different experts can naturally align with distinct spatial structures, providing scientists with an overview of how model capacity is distributed across the field.
\par
Furthermore, we introduce a parameter-conditioned adapter network that modulates each retrieved feature vector based on the simulation parameters, explicitly modeling how local field values should change under different input conditions. This design allows the expert assignments to be stable across the simulation parameter space, which facilitates targeted parameter-space exploration. For example, scientists can first identify a region of interest through the expert assignments and then vary the simulation parameters to analyze localized sensitivity within that region. 
\par
We evaluate \ours on three ensemble simulation datasets spanning oceanography, cosmology, and fluid dynamics. Beyond quantitative and qualitative evaluations, our study also emphasizes how the proposed framework can support scientific exploration through expert-guided exploration, interpretable expert specialization, and localized sensitivity analysis.
\par
In summary, our main contributions are as follows:
\begin{itemize}
    \item Proposing an adaptive INR framework with cross-attention that enables high-fidelity surrogate modeling for ensemble simulations while maintaining a compact model size.
    \item Introducing a coordinate-guided MoE architecture that not only improves feature utilization but also produces an interpretable spatial decomposition over the simulation domain.
    \item Supporting the expert-guided spatial analysis and targeted exploration of ensemble simulations through a two-stage workflow.
\end{itemize}

\section{Related Work}

\subsection{Implicit Neural Representations}

INRs have emerged as a powerful framework for learning continuous, memory-efficient representations of complex signals. The traditional approaches~\cite{mildenhall2021nerf, saragadam2023wire, sitzmann2020implicit, tancik2021learned, tancik2020fourier} employ multilayer perceptrons (MLPs) as the backbone to map input coordinates directly to the output signal values. 
However, fully implicit MLP-based models suffer from a well-known spectral bias toward low-frequency components, making it difficult to capture high-frequency details such as edges and fine-grained structures~\cite{rahaman2019spectral}.
Techniques, such as positional encodings and specialized activation functions~\cite{sitzmann2020implicit, mildenhall2021nerf}, have been introduced to mitigate this problem, while another line of work augments INRs with explicit data structures such as feature grids and planes~\cite{cao2023hexplane, fridovich2023k, fridovich2022plenoxels, muller2022instant, sun2022direct, sun2022improved}. These hybrid architectures substantially reduce training time and enhance the reconstruction quality, but often at the cost of increased memory usage and reduced model compactness.

\subsection{Mixture-of-Experts for INRs}

Recently, many INR methods have adopted the Mixture-of-Experts (MoE) architecture. In general, MoE consists of two major components: a router network that assigns the coordinate inputs to different sub-networks, and the expert sub-network that specializes in modeling specific patterns within the scene.  
Specifically, for papers on neural radiance fields (\ie, NeRF)~\cite{mildenhall2021nerf}, the goal of integrating MoE into the NeRF~\cite{di2024boost, rebain2021derf} model is mainly to improve the rendering quality for large-scale scenes and increase the model capacity without largely introducing the computational complexity during inference time. For example, Boost-Your-NeRF~\cite{di2024boost} proposes a model-agnostic MoE framework that models the scene at different spatial resolutions, allowing each expert to focus on regions at different frequencies. Switch-NeRF introduces a universal approach for large-scale scene decomposition. Unlike previous methods that first decompose the scene using data-driven heuristics and then train separate reconstruction models, it leverages the MoE's gating network to learn the decomposition and optimize the sub-networks during training directly.
\par
In addition to the MoE-based NeRFs, MoE has also been employed for general INR tasks, where multiple networks are used to increase the representational power and enable local approximation of continuous functions. 
In contrast, we carefully integrate Mixture-of-Experts (MoE) with key-value memory banks, allowing the model to allocate the augmented memory efficiently and utilize it effectively, which is sharply different from other MoE-based INRs that primarily focus on scalability~\cite{han2025moe, ben2024neural}. 
In the context of scientific simulation, a recent work~\cite{han2025moe} on data compression for time-varying volumetric data also leverages the MoE architecture. However, we are fundamentally different in several aspects. First, MoE-INR applies MoE to the decoder part, which routes the output of a shared encoder to the specialized decoder experts, while our method applies MoE to the spatial encoder part, where each expert maintains its own memory bank for spatially specialized feature encoding. Second, MoE-INR is entirely based on MLPs, whereas we proposed a novel design that leverages cross-attention for feature retrieval and utilizes a low-resolution feature grid for efficient spatial routing. Third, MoE-INR requires a pretraining stage to initialize the expert assignments and employs the hard Top-1 routing, in which each coordinate is assigned to a single expert based on pre-clustered voxel values. In contrast, our approach does not require any pretraining and uses the Top-2 soft routing. Furthermore, beyond the aforementioned technical differences, our encoder experts also provide interpretable spatial decomposition and localized parameter-space analysis for the ensemble exploration.

\subsection{Surrogate Models for Ensemble Simulations}

Exploring the physical phenomenon across varying simulation conditions is often computationally expensive \cite{chartier2021carpool, ohana2024well}. To mitigate this problem, numerous deep learning-based surrogate models have been proposed to accelerate simulations at significantly reduced cost while preserving fidelity.
For example, several studies focus on spatio-temporal forecasting \cite{ruhling2023dyffusion, lienen2022learning, takamoto2023learning, li2020fourier}, learning to approximate dynamic systems across both time and space.
For mesh-based simulations defined on irregular, unstructured grids, graph- or mesh-aware models such as MeshGraphNets \cite{pfaff2020learning} and GNN-Surrogate \cite{shi2022gnn} have been developed. 
Some other works employed models like autoencoders for surrogate modeling of volumetric ensemble simulations \cite{shi2022vdl, shen2024surroflow}.
However, most approaches are tied to discretized input spaces and designed to predict the entire discretized fields as outputs. As a result, they cannot support sparse querying and face the scalability limitations due to their high memory consumption. Moreover, their architectures are often tailored for specific types of simulations or domains, limiting their adaptability to broader scientific tasks.
\par
In addition to these grid- and mesh-based surrogate models, Neural Operators have emerged as another line of surrogate models that learn mappings between function spaces. Classical Neural Operators such as DeepONet~\cite{lu2019deeponet} and the Fourier Neural Operator~\cite{li2020fourier} approximate the underlying PDE solution operator by taking entire input functions, such as initial conditions or boundary conditions, and predicting the corresponding solution fields. 
Recent work has extended Neural Operators with transformer-based architectures \cite{li2022transformer,hao2023gnot,wen2025geometry,ovadia2024vito,wang2024cvit} to better handle irregular geometries and multi-scale problems through self-attention and cross-attention mechanisms. Despite their expressiveness, these methods still rely on fully discretized representations of both input and output functions, resulting in significant memory and computational cost, particularly for high-resolution 3D domains. Thus, such models may not be well-suited for dense 3D ensemble surrogate modeling or applications that require sparse coordinate-based querying at arbitrary resolutions.
\par
In contrast, INR-based surrogates take coordinates as inputs and directly learn continuous mappings to the field values, providing a compact and efficient framework for various scientific applications, including simulation surrogates~\cite{serrano2023infinity, catalani2024neural, chen2025explorable}, forecasting~\cite{yin2022continuous, luo2024continuous}, and reduced-order modeling of PDEs~\cite{chen2022crom}. 
Recently, the INR-based surrogate models have demonstrated strong performance in diverse scientific domains, ranging from fluid dynamics and climate simulations~\cite{pan2023neural, luo2024continuous} to medical imaging~\cite{shen2022nerp, song2023piner, reed2021dynamic}.
By learning grid- and mesh-agnostic representations, INRs {eliminate discretization constraints, reduce memory requirements, and enable flexible querying of continuous values at arbitrary coordinates or physical conditions}. 
Furthermore, INRs augmented with advanced architectural designs are particularly effective in capturing fine-scale and high-frequency structures, and have been shown to outperform domain-specific surrogates in applications such as ocean simulation~\cite{chen2025explorable}.

\begin{figure*}[htbp]
  \centering
  \includegraphics[width=0.99\textwidth]{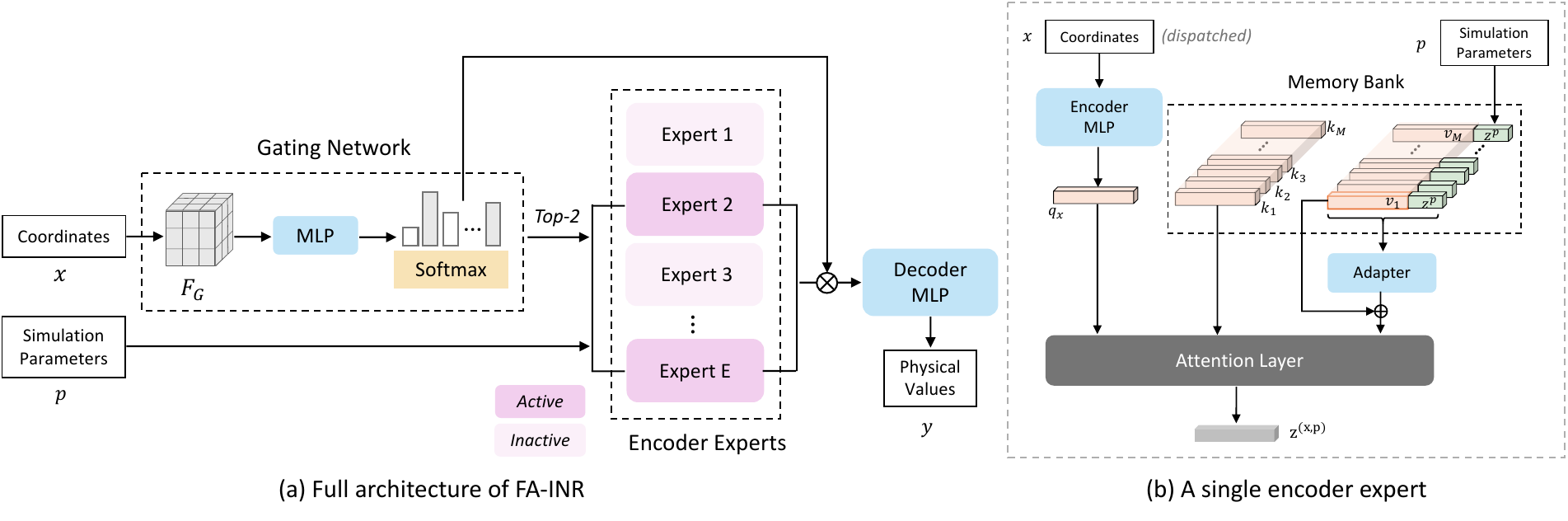}
  \caption{Architecture of the proposed \ours. An input pair $(x,p)$ is first routed by a gating network in (a) to the Top-2 relevant spatially-specialized encoder experts (see (b)). Then, the aggregated feature vector from all selected experts is further processed by the MLP decoder in (b).}
  \label{fig:fa_inr_framework}
\end{figure*}

\section{Methodology}

\subsection{Problem Definition}
\label{sec:problem_def}

We consider surrogate modeling for scientific ensemble simulations, where a numerical simulator generates physical fields over a predefined set of spatial coordinates, \ie, $X=\{x_1,x_2,...,x_N\}$, with each coordinate $x_i \in \mathbb{R}^{d}$.  
For a simulation parameters $p_j \in \mathbb{R}^{m}$, the simulator outputs a corresponding field $Y_j= \{y_{j,1}, y_{j,2},..., y_{j,N}\}$. Across the ensemble members, the spatial coordinates remain fixed while only the simulation parameters $p_j$ vary.
The objective of this surrogate model is to learn an approximate mapping function $g: \mathbb{R}^{N\times d} \times \mathbb{R}^{m} \rightarrow \mathbb{R}^{N}$, such that given the simulation parameters $p_j$, it can predict the corresponding output field $Y_j$ significantly faster than the original simulators.
\par
An INR aims to learn this mapping using a neural network $f_\theta$, parameterized by weights $\theta$. Specifically, the network maps an input pair ($x_i, p_j$) to its target output $y_{j,i}$, \ie, $f_\theta: \mathbb{R}^{d} \times \mathbb{R}^{m} \rightarrow \mathbb{R}$ such that $f_\theta(x_i, p_j) \approx y_{j,i}$.
Unlike other deep learning-based surrogate models that predict an entire discretized field $Y_j$ at once, this coordinate-based formulation enables resolution-independent querying at arbitrary spatial locations and simulation conditions.
\par
Given a training dataset $\sD=\{(X,p_1,Y_1),(X,p_2,Y_2),...,(X,p_J,Y_J)\}$ with $J$ simulation runs, the INR model $f_\theta$ can be trained by minimizing:
\begin{equation}
\mathcal{L}(\theta) = \frac{1}{J \cdot N} \sum_{j=1}^{J} \sum_{i=1}^{N} \ell(f_\theta(x_i, p_j), y_{j,i}),
\label{eq:inr_objective}
\end{equation}
where $\ell(\cdot)$ denotes a reconstruction loss such as mean squared error (MSE), which measures the differences between the predicted value $f_\theta(x_i, p_j)$ and the ground-truth value $y_{j,i}$.

\subsection{Network Architecture of \ours}
\label{method_sec}

Overall, \ours consists of an MoE-based feature encoder and an MLP-based feature decoder, as illustrated in \autoref{fig:fa_inr_framework}. The feature encoder contains three major components: a set of encoder experts, a gating network, and the parameter-conditioned feature adapter. 
\par
Given an input coordinate, the gating network assigns it to a specific expert according to its spatial location. Each encoder expert maintains its own key-value memory bank, which is shared by all coordinates routed to that expert. To model the parameter-dependent variations, the feature adapter takes the simulation parameters as input and modulates the retrieved feature vector accordingly. The final feature representation is then passed to the decoder for predicting the target field value. We provide the details of each component in the following subsections.

\subsubsection{Adaptive Feature Encoding}
\label{sec:cross_attention}

A standard MLP-based INR offers a compact continuous representation but often struggles to represent high-frequency variations~\cite{fridovich2022spectral, rahaman2019spectral}. Prior work addresses this by augmenting INRs with explicit feature encoding~\cite{cao2023hexplane, fridovich2023k, fridovich2022plenoxels, sun2022direct, sun2022improved} such as feature grids that store learnable feature vectors at discrete locations (vertices). While these structures are efficient, they allocate capacity uniformly across the 3D space and often require dense representations to capture complex details at high fidelity, resulting in substantial memory usage. This dense storage is often redundant, particularly for large-scale scientific datasets, where smooth regions may contain nearly uniform values. In contrast, in the rapidly varying areas, these rigid structures may under-represent the underlying data complexity.
\par
To overcome this limitation, FA-INR replaces rigid feature structures with a learnable key-value memory bank. The memory bank is defined as $\{K \in \mathbb{R}^{M\times{D_k}}, V \in \mathbb{R}^{M\times{D_v}}\}$, where $K$ denotes a set of key vectors representing feature locations, and $V$ denotes the corresponding feature vectors. Both $K$ and $V$ are learned during training, allowing the feature locations to adapt to the underlying data characteristics rather than being fixed at predefined positions. In contrast to fixed grid interpolation, this mechanism enables feature retrieval that dynamically adapts to local spatial variations.
\par
Specifically, to determine what features should be retrieved for a queried location, we leverage the cross-attention mechanism. As illustrated in \autoref{fig:fa_inr_framework}-(b), each query vector $q^{(x_i)}$ is produced from a spatial coordinate $x_i$ using an encoder MLP $f_{\theta_\text{E}}$. The encoder first maps $x_i$ to an intermediate feature vector ${z}^{(x_i)}$, which is then projected into the query space through a linear transformation.  Formally, this encoding process is defined as:
\begin{gather}
    q^{(x_i)} = {z}^{(x_i)}{W_q}, \hspace{4pt}  
    W_q \in \mathbb{R}^{D_q \times D_k}, 
    \\
    \text{where } {z}^{(x_i)} = f_{\theta_\text{E}}(x_i).
\end{gather}
Here, $\theta_\text{E}$ denotes the weights of the encoder MLP, and $W_q$ is a learnable projection matrix that maps the intermediate spatial features to the query dimension $D_k$.
\par
Each spatial query then retrieves its corresponding feature representation from the memory bank through cross-attention:
\begin{gather}
    z^{(x_i,p_j)} = \text{Softmax}(\frac{q^{(x_i)} (K W_k)^\top}{\sqrt{D_k}}) V^{(p_j)} W_v, 
    \label{eq:retrieval}
\end{gather}
where $W_k \in \mathbb{R}^{D_k \times D_k}$ and $W_v \in \mathbb{R}^{D_v \times D_v}$ are learnable linear projection matrices applied to keys and values, respectively.
In particular, $W_k$ follows the standard attention design, where a separate linear projection is learned to improve representational flexibility.
$V^{(p_j)}$ represents the learned feature embeddings after conditioning on the simulation parameter $p_j \in \mathbb{R}^{m}$. Details of this conditioning process are provided in the next paragraph.

\subsubsection{Parameter-Conditioned Feature Adaptation}
A surrogate model for ensemble simulations is expected to model not only spatial variations but also how the physical field changes under different simulation parameters. To this end, we present a parameter-efficient approach to transform the learned embedding $V$ to $V^{(p_j)}$ based on the simulation parameters $p_j$, as illustrated in \autoref{fig:fa_inr_framework}-(b).
Given $p_j \in \mathbb{R}^{m}$, we first embed each dimension $s$ ($1\leq s\leq m$) separately into a feature vector to characterize each simulation variable. 
These $m$ embeddings are then combined via an element-wise (Hadamard) product to form an aggregated embedding ${z}^{(p_j)}$. As mentioned in~\cite{fridovich2023k}, this operation better preserves the signal from individual variables. This aggregated embedding ${z}^{(p_j)}$ is then concatenated with each of the $M$ learned embeddings in $V \in \mathbb{R}^{M\times{D_v}}$ (\ie, the memory bank) and processed by a small MLP adapter $f_{\theta_\text{A}}$ to generate $V^{(p_j)}$:
\begin{gather}
    V^{(p_j)} = V + f_{\theta_\text{A}}({V \oplus z^{(p_j)}}) \in \mathbb{R}^{M\times{D_v}}.
\end{gather}
This residual connection is designed to better preserve the original information in $V$ while explicitly adapting the features based on the specific simulation parameters $p_j$.

\subsubsection{Spatially-Specialized Encoder Experts}
\label{sec:moe}

\begin{figure}[htbp]
  \centering
  \includegraphics[width=0.68\columnwidth]{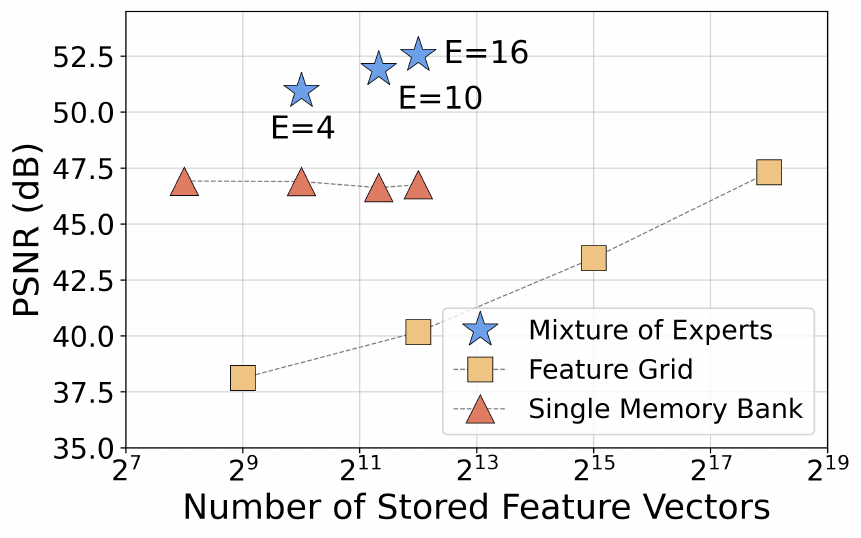}
  \caption{A single memory bank already outperforms the feature grid under the same feature storage budget. By introducing multiple encoder experts (MoE), our method further scales performance more effectively than simply expanding the memory bank, while requiring fewer feature vectors compared to the grid-based approach.}
  \label{fig:compare_grid_bank}
\end{figure}

Although a single attention-based memory bank is more flexible and data-adaptive than rigid grids, it must cover the entire spatial domain. For large-scale scientific fields, it is challenging for the key–value pairs in the memory bank to maintain effective utilization across all regions. Unlike feature grids, which are defined with spatial priors, here, both keys and values are randomly initialized. 
As illustrated in \autoref{fig:compare_grid_bank}, our approach with a relatively small memory bank can already outperform grid-based models. However, simply increasing the size of the memory bank results in limited performance improvements
\par
\begin{figure}[htbp]
  \centering
  \includegraphics[width=0.80\columnwidth]{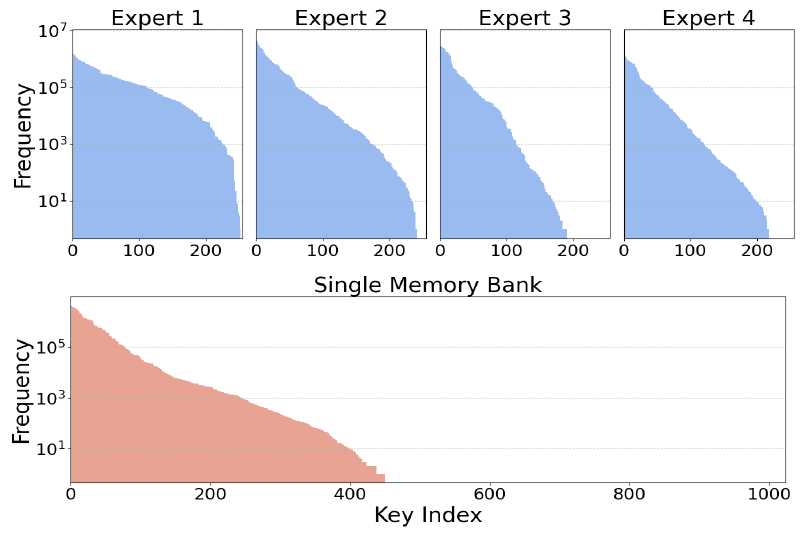}
  \caption{Key usage comparison between a single memory bank and the MoE design. In this example, the MoE setup with four experts distributes queries more evenly across keys within each expert, resulting in more effective key utilization.}
  \label{fig:key_hists}
\end{figure}

To enhance the key-value pair usage, we introduce an additional constraint that organizes these pairs into specialized groups and use the spatial coordinate $x_i$ as the routing signal to determine which group should be queried. This strategy naturally aligns with the idea of the Mixture-of-Experts (MoE) framework.
As a result, the keys are more efficiently utilized within each group, as demonstrated in \autoref{fig:key_hists}.


Formally, we introduce our coordinate-guided MoE architecture in which each expert is an attention-based feature encoder with its own memory bank (see \autoref{fig:fa_inr_framework}-(a)). To dynamically select the most relevant experts for each coordinate $x_i$, we design a gating network consisting of two components: a small, low-resolution feature grid $\textit{F}_\text{G}$ that provides a coarse spatial representation, and a lightweight MLP $f_{\theta_\text{G}}$ parameterized by $\theta_\text{G}$. 
\par
Given an input coordinate $x_i$, a coarse spatial feature ${z_\text{G}}^{(x_i)}$ is first extracted from $\textit{F}_\text{G}$. This feature vector is then processed by the MLP $f_{\theta_\text{G}}$. Finally, a softmax function is applied to the output to produce the expert selection probabilities:
\begin{gather}
    \Phi(x_i) = \text{softmax}(f_{\theta_\text{G}}({z_\text{G}}^{(x_i)})), \hspace{4pt} \Phi(x_i) \in \mathbb{R}^{E},
    \label{eq:gating_probabilities}
\end{gather}
where $E$ denotes the total number of expert modules, and $\Phi(x_i)$ is the probability distribution over these experts.
\par
Given the gating probabilities $\Phi(x_i)$, the Top-$K$ encoder experts are selected for each coordinate $x_i$. The final aggregated feature ${\bar z}^{(x_i,p_j)}$ is then computed as a weighted sum of the feature vectors ${z_k}^{(x_i,p_j)}$ produced by the selected experts:
\begin{gather}
    {\bar z}^{(x_i,p_j)} = \sum_{k=1}^{K} {\Phi(x_i)}_k \cdot {z_k}^{(x_i,p_j)},
    \label{eq:moe_aggregation} 
\end{gather}
where ${\Phi(x_i)}_k$ is the probability of routing $x_i$ to the $k$-th expert, and ${z_k}^{(x_i,p_j)}$ represents the corresponding output feature vector from the $k$-th expert encoder, conditioned on both the spatial coordinate $x_i$ and the simulation parameter $p_j$.

\subsubsection{Feature Decoder}
\label{sec:decoder}

To decode the aggregated feature vector ${\bar z}^{(x_i,p_j)}$ into a physical value, a small MLP is employed as the feature decoder $f_{\theta_\text{D}}$, as shown in \autoref{fig:fa_inr_framework}-(b). This decoder maps the feature representation to the predicted value $\hat{y}_{j,i}$:
\begin{gather}
    \hat{y}_{j,i} = f_{\theta_\text{D}}({\bar z}^{(x_i,p_j)}).
\end{gather}

\subsection{Optimization}
\label{sec:optimization}

The entire framework, including all encoder experts, the gating network, the parameter-conditioned feature adapter, and the decoder, is trained end-to-end. We optimize the model by minimizing the mean squared error (MSE) between the predicted scalar values $\hat{y}_{j,i}$ and the ground-truth values $y_{j,i}$, across all ensemble members and sampled spatial coordinates.
The training objective is formulated as follows:
\begin{gather}
    \mathcal{L}_{\text{MSE}}(\theta) 
    = \frac{1}{{J \cdot N}}\sum_{j=1}^{J}\sum_{i=1}^{N}\left\|f_{\theta_\text{D}}({\bar z}^{(x_i,p_j)}) - y_{j,i}\right\|_2^2, 
    \label{eq:loss_mse}
\end{gather}
where $\theta$ represents all learnable parameters of the model, and $y_{j,i}$ is the ground-truth scalar value at coordinate $x_i$ for the $j$-th simulation run in the ensemble dataset.



\section{Results}
\label{sec:results}

We evaluate \ours on three ensemble simulation datasets. In \cref{sec:quantitative_eval} and \cref{sec:qualitative_eval}, we compare our method against four baselines, including both INR-based surrogate models and state-of-the-art INR methods. In \cref{sec:exploration}, we demonstrate how scientists can benefit from our expert-guided exploratory workflow.

\begin{table}[htbp]
\centering
\footnotesize
\setlength{\tabcolsep}{6pt}
\caption{Dataset statistics and training setups.}
\label{table:data_info}
\begin{tabular}{l|c|c|c|c}
Dataset & \makecell{Simulation\\parameters} & Coordinates & \makecell{Train\\members} & \makecell{Test\\members} \\
\hline
MPAS-Ocean   & 4 & $11{,}845{,}146$ & 70  & 30  \\
Nyx          & 3 & $512^3$          & 100 & 30  \\
CloverLeaf3D & 6 & $128^3$          & 500 & 100 \\
\end{tabular}
\end{table}

\subsection{Experimental Setup}
\label{sec:exp_setup}

\noindent \textbf{Datasets.} 
We evaluate our model and baselines on three ensemble simulation benchmarks: one unstructured ocean simulation and two structured volumetric simulations.
\par
\textit{MPAS-Ocean}~\cite{ringler2013multi} is a global ocean simulation defined in spherical coordinates (\ie, latitude, longitude, and depth) that models ocean temperature fields. This simulation is controlled by four parameters: Bulk Wind Stress Amplification (\textit{BwsA}), Critical Bulk Richardson Number (\textit{CbrN}), Gent--McWilliams eddy transport coefficient (\textit{GM}), and Horizontal Viscosity (\textit{HV}). This dataset contains total 100 simulation runs, each corresponding to a unique combination of these four parameters. We use 70 instances for training and the remaining 30 for testing.
\par
\textit{Nyx}~\cite{almgren2013nyx} is a cosmological hydrodynamics simulation that models the dark matter density fields. This ensemble dataset is generated by varying three cosmological parameters: the total density of baryons (\textit{OmB}),  the total matter density (\textit{OmM}), and the Hubble Constant (\textit{h}). We use 100 instances for training and 30 instances for testing.
\par
\textit{CloverLeaf3D} is a hydrodynamics simulation dataset generated by solving the 3D compressible Euler equations. This simulation is parameterized by six variables, including three density parameters and three energy parameters. In our experiments, we randomly sample 500 instances for training and 100 instances for testing.
\par
Additional dataset statistics, including the number of ensemble members, number of spatial coordinates, and number of simulation variables, are summarized in \autoref{table:data_info}.

\indent \textbf{Baselines.}
We compare \ours against four baseline approaches:
(1) Explorable-INR~\cite{chen2025explorable}, an INR-based surrogate model, combines a 3D feature grid with multiple 2D planes to balance model efficiency and accuracy. Since Explorable-INR has already been shown to outperform several closely related non-INR surrogate models (\ie, VDL-Surrogate~\cite{shi2022vdl}, GNN-Surrogate~\cite{shi2022gnn}, and InSituNet~\cite{he2019insitunet}), we adopt it as a strong baseline model for comparison.
(2) MMGN~\cite{luo2024continuous} is an INR-based surrogate model that employs a context-aware indexing mechanism together with a set of multiplicative basis functions to reconstruct high-quality physical fields from sparse observations.
(3) Neural Experts~\cite{ben2024neural} extends the MLP-based SIREN~\cite{sitzmann2020implicit} by incorporating a MoE architecture to learn local, piecewise continuous approximations.
(4) K-Planes~\cite{fridovich2023k} represents arbitrary-dimensional data by factorizing it into multiple 2D feature planes, providing a flexible and efficient surrogate modeling approach. We do not include MoE-INR~\cite{han2025moe} in our comparison mainly because its official implementation is not publicly available.
\par
\indent \textbf{Evaluation Metrics.}
We evaluate the performance of all methods using three metrics. Peak Signal-to-Noise Ratio (PSNR)~\cite{huynh2008scope} quantifies the numerical fidelity of predicted scalar fields with respect to the ground-truth simulations, where higher PSNR indicates more accurate approximations of the true signals. To capture the worst-case deviations, we compute the normalized maximum difference (MD)~\cite{shi2022gnn, chen2025explorable}, which estimates the relative upper bound of the prediction error. Finally, to assess perceptual quality, we measure the Structural Similarity Index (SSIM)~\cite{wang2004image} between the ground-truth and predicted fields. To ensure a fair comparison, the rendered results of all methods are generated using identical viewpoints and rendering configurations.
\par
\indent \textbf{Experimental Details.}
All models are implemented in PyTorch and trained on the NVIDIA A100 GPU. Moreover, all methods are optimized using the Adam optimizer with a consistent batch size to ensure fair comparisons. The initial learning rate is set to $1 \times 10^{-4}$ and decayed by 10\% during training. For Neural Experts, which uses a fully MLP-based architecture, we employ a smaller learning rate of $1 \times 10^{-6}$ on the Nyx dataset to ensure stable training.
\par
For our model, we use a consistent configuration for MPAS-Ocean and CloverLeaf3D datasets, while slightly increasing the model capacity for Nyx. Specifically, we use 256 key-value pairs in the memory bank for MPAS-Ocean and CloverLeaf3D, and 1,024 pairs for Nyx. For the MoE component in \ours, we follow the standard sparsely-gated design~\cite{shazeer2017outrageously} with Top-2 expert routing.
\par
To ensure fair comparisons, we slightly increase the model capacity of Neural Experts and MMGN by enlarging their network depth and hidden dimensions to match the number of trainable parameters of our model. This adjustment is necessary because both methods were originally designed for relatively simpler settings, such as 2D data. For Explorable-INR, we retain its original architecture and set the dimensions of our spatial and simulation-parameter embeddings to match those used in its model.

\begin{table}[htbp]
\centering
\footnotesize
\setlength{\tabcolsep}{4pt}
\caption{Quantitative comparison of \ours with baseline methods on the MPAS-Ocean dataset.}
\label{tab:main_results_ocean}
\begin{tabular}{l|c|c|c|c|c}
Method & \#Experts & Size (MB) & PSNR (dB)$\uparrow$ & MD$\downarrow$ & SSIM$\uparrow$ \\
\hline
FA-INR (Ours)   & 10 & 4.19   & \underline{\textbf{51.72}} & 0.157 & \underline{\textbf{0.993}} \\
Explorable-INR  & -  & 28.16  & 49.49 & 0.184 & 0.990 \\
K-Planes        & -  & 164.68 & 44.77 & \underline{\textbf{0.155}} & 0.989 \\
Neural Experts  & 10 & 4.42   & 41.01 & 0.334 & 0.981 \\
MMGN            & -  & 4.40   & 42.99 & 0.244 & 0.985 \\
\end{tabular}
\end{table}

\begin{table}[htbp]
\centering
\footnotesize
\setlength{\tabcolsep}{4pt}
\caption{Quantitative comparison between \ours and four baselines on the Nyx and CloverLeaf3D datasets.}
\label{tab:main_results_volumn}
\begin{tabular}{l|c|c|c|c|c}
Method & \#Experts & Size (MB) & PSNR (dB)$\uparrow$ & MD$\downarrow$ & SSIM$\uparrow$ \\
\hline
\multicolumn{6}{c}{Nyx}\\
\hline
FA-INR (Ours)   & 8 & 36.82  & \underline{\textbf{44.70}} & \underline{\textbf{0.081}} & \underline{\textbf{0.957}} \\
Explorable-INR  & - & 56.19  & 43.09 & 0.117 & 0.943 \\
K-Planes        & - & 154.97 & 35.32 & 0.156 & 0.894 \\
Neural Experts  & 8 & 35.12  & 38.17 & 0.171 & 0.924 \\
MMGN            & - & 33.50  & 39.77 & 0.205 & 0.925 \\
\hline
\multicolumn{6}{c}{CloverLeaf3D}\\
\hline
FA-INR (Ours)   & 10 & 4.19  & \underline{\textbf{53.40}} & \underline{\textbf{0.051}} & \underline{\textbf{0.967}} \\
Explorable-INR  & -  & 28.16 & 48.90 & 0.055 & 0.946 \\
K-Planes        & -  & 185.00 & 46.35 & 0.064 & 0.940 \\
Neural Experts  & 10 & 4.42  & 52.40 & 0.063 & 0.966 \\
MMGN            & -  & 4.40  & 45.52 & 0.069 & 0.931 \\
\end{tabular}
\end{table}


\subsection{Quantitative Evaluation}
\label{sec:quantitative_eval}

\indent \textbf{Surrogate Prediction.}
For the unstructured MPAS-Ocean dataset, \autoref{tab:main_results_ocean} summarizes the performance of all methods in approximating the simulation outputs under {unseen simulation parameters}. Our method, \ours, consistently outperforms all baselines, achieving an average improvement of 7.16 dB in PSNR.
Moreover, \ours attains the lowest MD, demonstrating its ability to suppress large prediction errors. This robustness is particularly important for reliable and efficient parameter-space exploration in scientific applications.
While Explorable-INR achieves competitive performance, it relies on a combination of one feature grid and three high-resolution planes, resulting in a model with approximately 7$\times$ more parameters than ours.
\par
Compared to the global ocean simulation, volumetric simulations such as Nyx and CloverLeaf3D often present denser and more complex 3D features. In addition, spatial patterns in these datasets can vary significantly across different simulation parameters. These characteristics introduce additional challenges for INR-based surrogate models, which require accurate modeling of fine-grained spatial details and parameter-dependent variations. 

The quantitative results for the volumetric datasets are summarized in \autoref{tab:main_results_volumn}. Overall, our \ours consistently achieves the highest performance across both datasets, with average improvements of 5.61 dB on Nyx and 5.11 dB on CloverLeaf3D in PSNR.
While Explorable-INR performs reasonably well on Nyx, its PSNR drops significantly on CloverLeaf (48.90 dB), indicating limited adaptability to ensemble data with highly variable spatial features. In contrast, Neural Experts (with MoE) achieves competitive performance on CloverLeaf3D (52.40 dB) but underperforms on Nyx, suggesting that employing MoE architecture alone is insufficient to handle the diverse spatial and parameter-dependent structures. The strong and consistent performance of \ours across both benchmarks highlights the effectiveness of our \textit{expert specialization} and \textit{adaptive feature encoding} designs.
\par
\begin{table}[htbp]
\centering
\footnotesize
\setlength{\tabcolsep}{9.8pt}
\caption{Evaluation of generalization to unseen spatial locations on the MPAS-Ocean dataset, where T denotes trained spatial locations, and U denotes unseen spatial locations.}
\label{tab:unseen_x}
\begin{tabular}{l|c|c|c}
Model & \#Experts & PSNR (T/U)$\uparrow$ & MD (T/U)$\downarrow$ \\
\hline
FA-INR (Ours)   & 10 & \textbf{52.20} / \textbf{51.43} & \textbf{0.165} / \textbf{0.173} \\
Explorable-INR  & -- & 49.89 / 49.15 & 0.391 / 0.391 \\
K-Planes        & -- & {51.04} / 43.38 & 0.166 / {0.173} \\
Neural Experts  & 10 & 40.28 / 39.60 & 0.413 / 0.404 \\
MMGN            & -- & 44.17 / 39.27 & 0.347 / 0.353 \\
\end{tabular}
\end{table}

\indent \textbf{Spatial Generalization.}
We further evaluate the generalization capability of \ours to unseen spatial locations using the MPAS-Ocean dataset. Although our method is primarily designed to improve the generalization of surrogate prediction across unseen simulation parameters, this additional evaluation examines its ability as an INR to naturally extend to unseen spatial coordinates.
Specifically, during training, we randomly sample 70\% of the spatial coordinates from each of the 70 ensemble members and reserve the remaining 30\% for testing. Under this setting, the simulation parameters are fixed, and the evaluation solely focuses on generalization to unseen locations.
\par
The quantitative results on MPAS-Ocean are summarized in \autoref{tab:unseen_x} (\ie, ``Trained'' columns). Overall, \ours consistently outperforms all INR-based baselines, achieving an average improvement of 5.86 dB in PSNR. Interestingly, Explorable-INR shows relatively limited spatial generalization compared to its performance in the surrogate prediction experiments. In contrast, K-Planes achieves competitive results on both metrics, suggesting that although it struggles to generalize across unseen simulation parameters, its multi-resolution design of feature plane is particularly effective at modeling spatial variations. However, this advantage comes at the cost of a significantly larger model size.

We further evaluate all models on \textit{unseen ensemble members}, which is a more challenging setting because both the spatial locations and the simulation parameters are unseen during training. As shown in \autoref{tab:unseen_x} (\ie, ``Unseen'' columns), \ours again achieves the best performance among all methods. K-Planes exhibits the largest performance drop from the ``Trained'' setting to the ``Unseen'' setting, indicating its limited ability to generalize to unseen simulation parameters.

\begin{table}[htbp]
\centering
\caption{Performance of \ours with different numbers of encoder experts.}
\footnotesize
\setlength{\tabcolsep}{8pt}
\label{tab:num_experts}
\begin{tabular}{l|c|c|c|c}
\multicolumn{5}{c}{{MPAS-Ocean}} \\
\hline
\#Experts & 1 & 4 & 10 & 16 \\
\hline
\#Trainable params. & 0.20M & 0.50M & 1.10M & 1.69M \\
PSNR (dB)$\uparrow$ & 46.92 & 50.94 & 51.92 & \textbf{52.55} \\
MD$\downarrow$ & 0.209 & 0.167 & \textbf{0.153} & 0.177 \\
\hline
\multicolumn{5}{c}{{Nyx}} \\
\hline
\#Experts & 1 & 8 & 10 & 12 \\
\hline
\#Trainable params. & 1.36M & 9.65M & 12.02M & 14.39M \\
PSNR (dB)$\uparrow$ & 39.49 & 44.72 & 45.63 & \textbf{45.67} \\
MD$\downarrow$ & 0.097 & 0.083 & \textbf{0.071} & 0.076 \\
\end{tabular}
\end{table}

\begin{table}[htbp]
\centering
\caption{Effect of feature grid resolution in the gating module on the MPAS-Ocean dataset.}
\footnotesize
\setlength{\tabcolsep}{12pt}
\label{tab:diff_gate_res}
    \begin{tabular}{l|c|c|c}
    Resolution & $16^3$ & $32^3$ & $64^3$ \\
    \hline
    \#Trainable params. & 1.10M & 1.56M & 5.23M \\
    \#Experts & 10 & 10 & 10 \\
    PSNR (dB)$\uparrow$ & 51.92  & 51.59  & 52.66 \\
    MD$\downarrow$ & 0.153 & 0.156 & 0.160 \\
    \end{tabular}
\end{table}


\indent \textbf{MoE Design Analysis.}
We investigate the impact of increasing the number of encoder experts on model performance. In our experiments, all experts share the same network architecture, and each spatial coordinate is routed to the Top-2 experts. As shown in \autoref{tab:num_experts}, increasing the number of experts significantly improves performance on the MPAS-Ocean dataset. However, the performance gain becomes marginal beyond 10 experts, and the MD metric slightly degrades. A similar trend is observed on the Nyx dataset: increasing the number of experts from 10 to 12 yields only a minor improvement in PSNR.
It is worth noting that the MoE framework in \ours is applied only to the feature encoding component rather than the entire INR architecture. Therefore, increasing the number of experts introduces only a small overhead in training time and model size.
\par
As introduced in \cref{sec:moe}, the gating module contains a low-resolution feature grid and a linear projection layer. In our main experiments, we set the feature grid resolution at $16^3$ for all three datasets, as a small feature grid provides more efficient and stable optimization than alternatives such as MLP-based gating, particularly during the early stages of training.
Here, we investigate how varying the feature grid resolution affects overall model performance. As shown in \autoref{tab:diff_gate_res}, a low-resolution grid ($16^3$) is sufficient to achieve high-fidelity results. Increasing the grid resolution does not necessarily improve performance, while the number of model parameters substantially increases. This observation is consistent with findings from previous studies~\cite{ben2024neural, di2024boost}.

\begin{figure}[b!]
    \centering
    \noindent\includegraphics[width=1.0\linewidth]{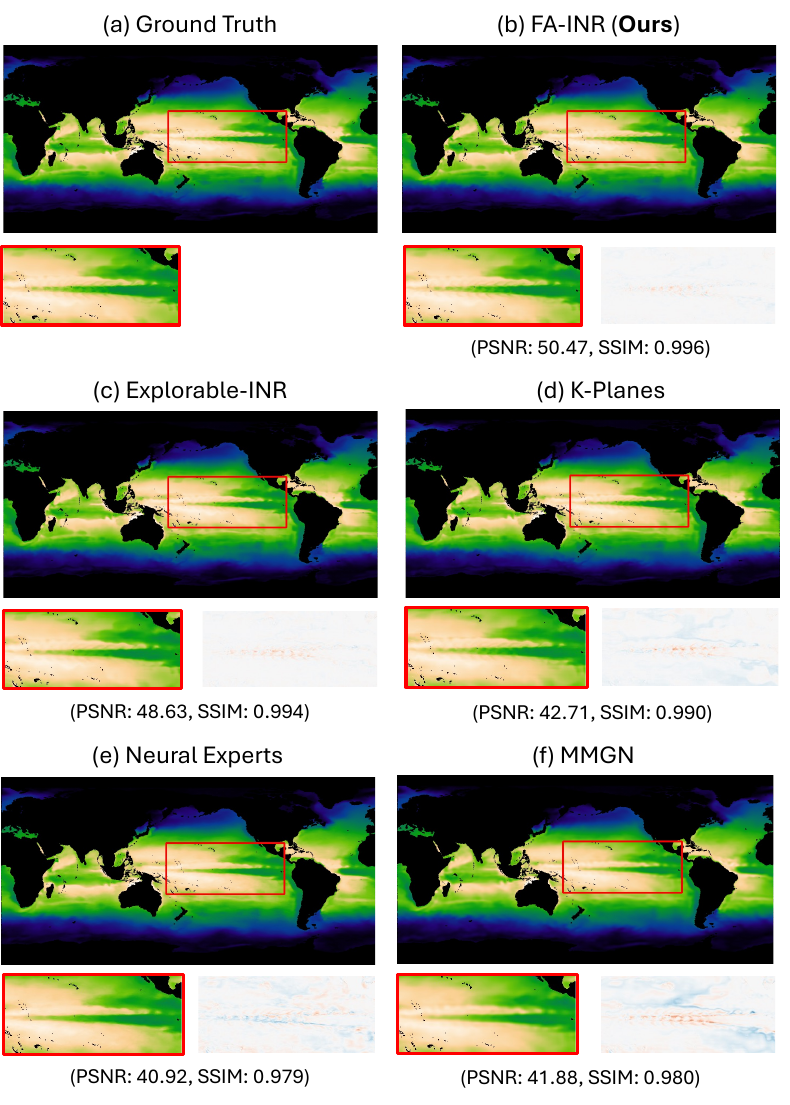}
    \caption{The comparison of visual fidelity on a test member of the MPAS-Ocean dataset.}
    \label{fig:rendering_mpaso}
\end{figure}

\begin{figure}[htbp]
    \centering
    \noindent\includegraphics[width=1.0\linewidth]{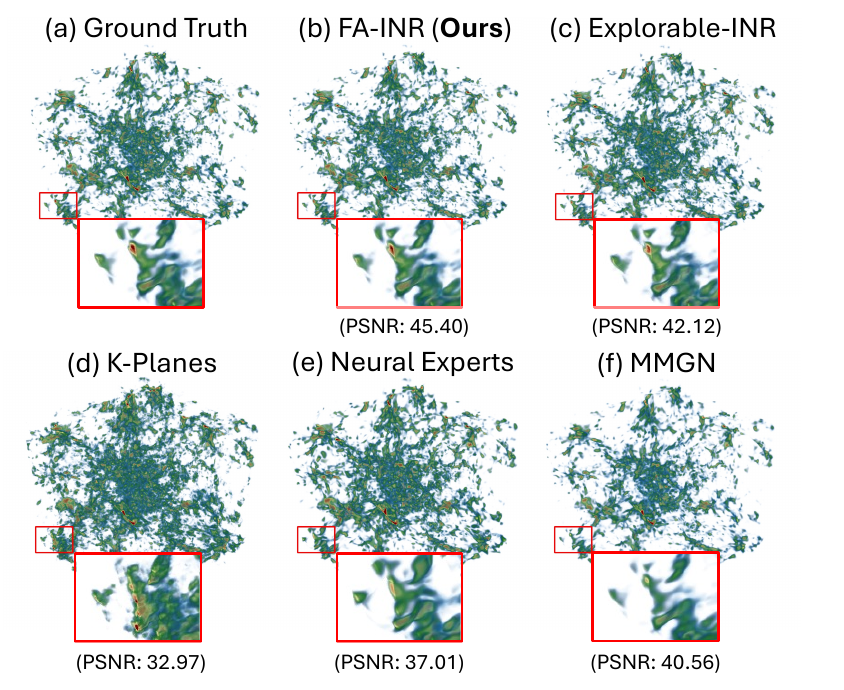}
    \caption{The comparison of visual fidelity on a test member of the Nyx dataset. The PSNR reported below each image denotes the reconstruction accuracy of the underlying scalar field.}
    \label{fig:rendering_nyx}
\end{figure}

\begin{figure}[t!]
    \centering
    \noindent\includegraphics[width=1.0\linewidth]{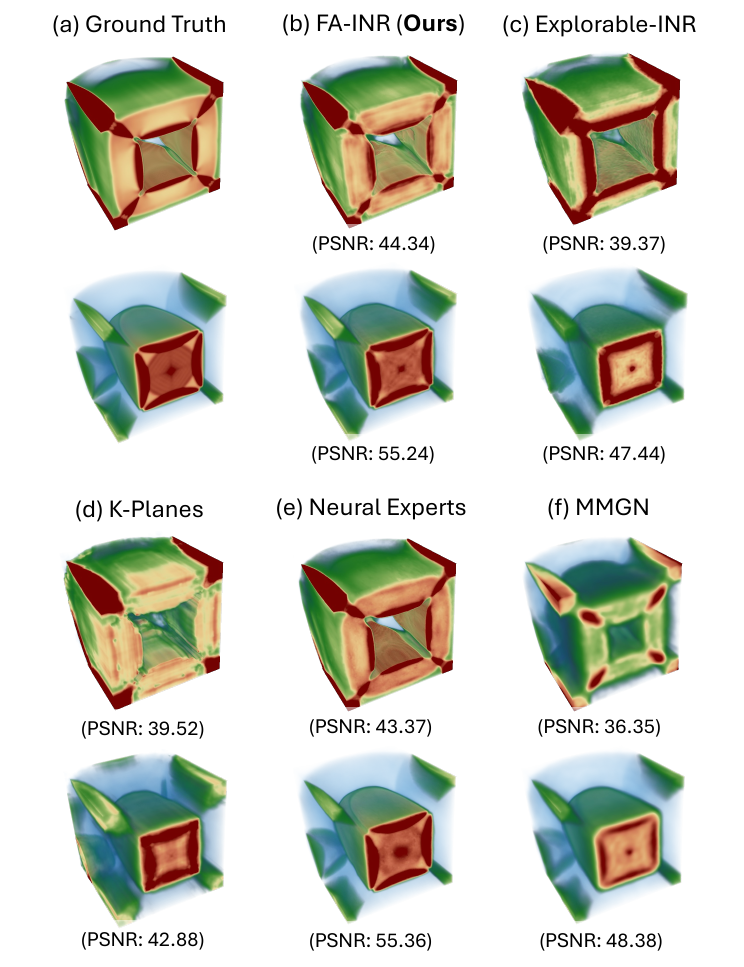}
    \caption{Comparison of visual fidelity on two representative test members of the CloverLeaf3D dataset. Large visual differences can be observed across the rendering results of all methods. Both \ours and Neural Experts, which employ the MoE architecture, achieve the highest prediction accuracy. The PSNR reported below each image denotes the reconstruction accuracy of the underlying scalar field.}
    \label{fig:rendering_cloverleaf3d}
\end{figure}

\subsection{Qualitative Evaluation}
\label{sec:qualitative_eval}

To demonstrate the high visual quality achieved by \ours, we report the SSIM scores across all approaches in both \autoref{tab:main_results_ocean} and \autoref{tab:main_results_volumn}. 
\par
For the MPAS-Ocean dataset, each ensemble member contains 60 depth levels. For visual comparison, we select the surface layer of the ocean, which presents rich and highly varying spatial features. All images are rendered using ParaView based on the Visualization Toolkit (VTK) libraries. 
In \autoref{fig:rendering_mpaso}, we compare the rendering results of our method with those of the baseline approaches. A difference map is provided alongside each zoomed-in visualization to highlight prediction errors. The cold tongue region in the difference maps demonstrates that \ours achieves the lowest reconstruction error and the highest visual fidelity.
\par
For the Nyx dataset, our method again achieves the highest visual fidelity. In contrast, as shown in the zoomed-in views of \autoref{fig:rendering_nyx}-(d), K-Planes presents the most noticeable visual degradation across all methods. In particular, K-Planes introduces numerous artifacts when modeling the small filamentary structures in the density field. These artifacts could be attributed to two reasons. First, the cosmology data contains highly complex 3D structures that are difficult to approximate using only 2D feature representations.
Second, the use of multi-resolution feature planes may further amplify those false patterns. Furthermore, Neural Experts fails to capture high-frequency details, as illustrated in \autoref{fig:rendering_nyx}-(e). This limitation likely comes from its underlying MLP-based architecture, which lacks sufficient capacity to accurately model scientific data with strong spatial variability.
\par
Finally, \autoref{fig:rendering_cloverleaf3d} presents the visual comparisons between \ours and the baseline approaches on the CloverLeaf3D dataset. We select two representative ensemble members for qualitative evaluation. Compared with the results from the other two datasets, larger visual differences are observed across all methods on this more challenging dataset. Both \ours and Neural Experts achieve the highest visual accuracy in these two instances. A possible explanation is that both methods leverage the Mixture-of-Experts (MoE) framework, which enables them to better handle the complex and highly varying spatial patterns present in CloverLeaf3D.

\section{Expert-Guided Exploration}
\label{sec:exploration}

\subsection{Two-Stage Workflow}

After training the surrogate model, scientists can use it to explore physical outputs under arbitrary simulation conditions. However, when exploring the large-scale ensemble simulations, scientists often face two key challenges. 
First, the spatial domain is often large and heterogeneous, making it difficult for scientists to quickly identify regions of interest.
Second, when sweeping across the simulation parameter space, different spatial regions often present varying levels of sensitivity. Existing works typically measure parameter sensitivity over the entire data field, which averages the responses across all regions and overlooks the localized behaviors. As a result, without spatially targeted analysis, it becomes difficult for scientists to determine how individual simulation parameters influence the critical locations.
\par
In this section, we demonstrate how \ours can address these challenges by enabling \textit{an expert-guided two-stage exploration workflow}. In the first stage, scientists examine the expert assignment map to understand which expert is responsible for each spatial location. In our method, expert assignments are produced through data-driven optimization rather than predefined spatial partitioning. Therefore, the resulting spatial coverage could naturally reflect the learned specialization of each expert with respect to the underlying data characteristics. By further performing the frequency analysis within each expert's spatial region, scientists can identify which experts specialize in modeling high-frequency structures.
\par
Once a region of interest is identified, such as regions containing highly complex structures or areas showing specific scientific phenomena, scientists can proceed to the second stage and conduct the spatially targeted parameter analysis. Specifically, by focusing on the selected expert, expert-specific metrics such as data variance or frequency measures can be computed to quantify how physical values within that region respond to the changes in simulation parameters. More importantly, parameter sensitivity can be evaluated directly within the selected region of interest, enabling localized sensitivity analysis to provide meaningful insights for scientific reasoning. In the following sections, we present two case studies on the MPAS-Ocean and CloverLeaf3D datasets to demonstrate the effectiveness of this workflow.

\begin{table}[htbp]
\centering
\caption{Quantitative summary of per-expert performance and frequency measure for a representative member within the MPAS-Ocean dataset.}
\footnotesize
\setlength{\tabcolsep}{8pt}
\label{tab:experts_ocean}
    \begin{tabular}{l|c|c|c|c|c}
    Expert ID & E1 & E2 & E3 & E4 & E5 \\
    \hline
    PSNR (dB) & 48.44 & 48.67 & \underline{53.01} & 46.74 & 45.02 \\
    Frequency & 0.027 & 0.021 & 0.006 & \underline{0.047} & 0.018 \\
    \end{tabular}
\end{table}

\begin{figure*}[htbp]
    \centering
    \noindent\includegraphics[width=0.99\linewidth]{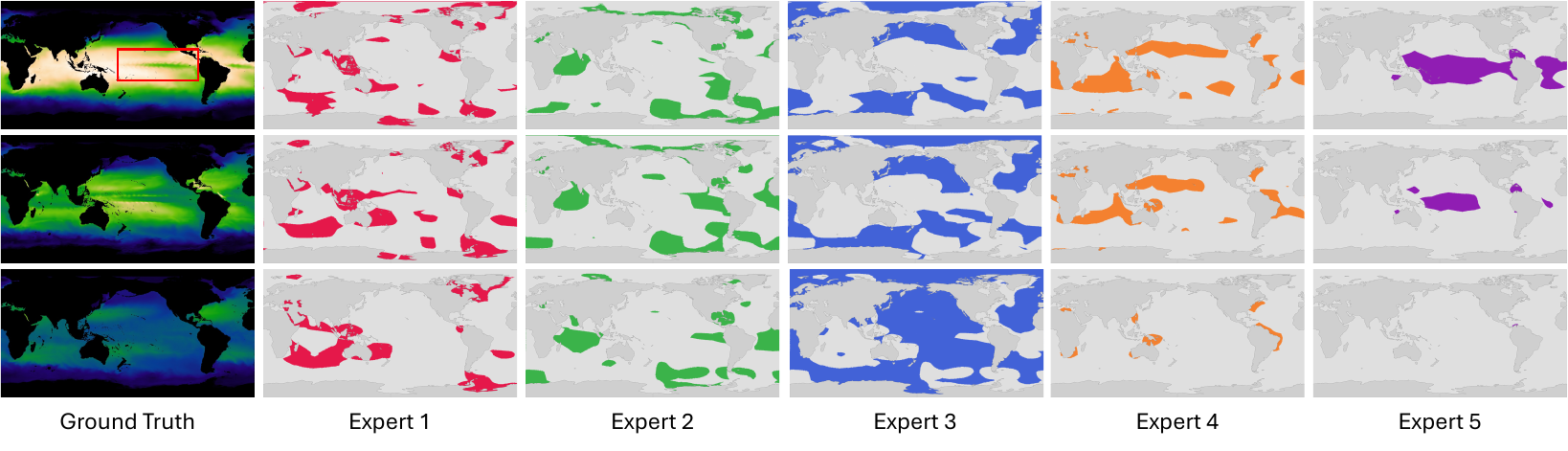}
    \caption{Expert assignment maps for a representative MPAS-Ocean ensemble member across three depth layers. The equatorial cold tongue is highlighted in the view of the ground-truth surface layer.}
    \label{fig:expert_assignment_ocean}
\end{figure*}

\subsection{Case Studies}

\subsubsection{Case Study with the MPAS-Ocean Simulation}

The global ocean simulation involves nearly 12 million spatial coordinates across 60 depth levels, making it time-consuming for the scientists to manually inspect the entire spatial domain. In this case study, we showcase how our spatially specialized experts can learn interpretable partitions and provide useful insights into the underlying scientific phenomena. 
\par
\autoref{fig:expert_assignment_ocean} presents the expert assignment maps at three selective ocean layers for a representative ensemble member. In addition, \autoref{tab:experts_ocean} summarizes the performance and frequency measure of each expert, where the frequency is computed using the graph Laplacian energy~\cite{shuman2013emerging} based on the ground-truth scalar field. Overall, Expert-3 achieves the highest PSNR but with the lowest frequency measure. The assignment maps indicate that Expert-3, covering the largest spatial region, is mainly responsible for modeling the large smooth features across the ocean layers. In contrast, Expert-4 focuses on modeling the ocean current boundaries, where the temperature field presents the highest-frequency variations. 

\begin{figure}[htbp]
    \centering
    \noindent\includegraphics[width=0.95\linewidth]{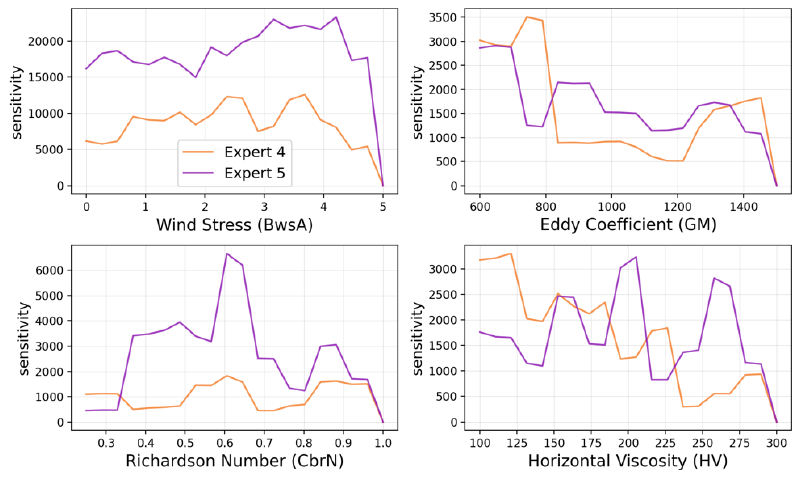}
    \caption{Parameter sensitivity of four simulation parameters in the MPAS-Ocean dataset for two selected experts.}
    \label{fig:E4E5_ocean}
\end{figure}
\par
Moreover, by comparing the ground-truth scalar field with each expert assignment map, we observe that the spatial region covered by Expert-5 aligns precisely with the eastern equatorial Pacific cold tongue (highlighted in red in \autoref{fig:expert_assignment_ocean}). The spatial coverage of Expert-4 gradually disappears in deeper ocean layers, further indicating that the learned experts indeed capture meaningful depth-dependent oceanographic structures.
\par
We then select Expert-4 and Expert-5 to further investigate the parameter sensitivity. This analysis aims to quantify how specific physical parameter impacts the simulation outputs. For each parameter sweep, the sensitivity is computed as the absolute gradient of the L1 norm of the predicted values with respect to that parameter~\cite{shi2022vdl,he2019insitunet}. The resulting sensitivity curves for both experts are presented in \autoref{fig:E4E5_ocean}.
Overall, \textit{wind stress (BwsA)} shows the strongest influence on the ocean temperature field. While Expert-4 and Expert-5 share the same ranking of parameter sensitivity, the influence of wind stress differs significantly between these two regions. This observation suggests that wind stress has a relatively stronger impact on the equatorial cold tongue compared to the ocean current boundaries. Furthermore, the sensitivity plot for \textit{CbrN} reveals a localized peak around 0.6 for Expert-4, while Expert-5 remains nearly unaffected. Without leveraging the expert specialization, such spatially localized behaviors would be difficult to identify using the conventional global parameter-space analysis.

\begin{table}[htbp]
\centering
\caption{Quantitative summary of per-expert performance and frequency measure, averaged across all test members of the CloverLeaf3D dataset.}
\footnotesize
\setlength{\tabcolsep}{2.2pt}
\label{tab:experts_cloverleaf}
    \begin{tabular}{l|c|c|c|c|c}
    Expert ID & E1 & E2 & E3 & E4 & E5 \\
    \hline
    PSNR (dB) & \underline{57.85} & 56.46 & 52.74 & 53.33 & 52.78 \\
    Frequency & $1.1\times10^{-3}$ & $3.0\times10^{-3}$ & $\underline{3.8\times10^{-3}}$ & $0.8\times10^{-3}$ & $0.8\times10^{-3}$ \\
    \end{tabular}
\end{table}

\subsubsection{Case Study with the CloverLeaf3D Simulation}

In the next case study, we demonstrate how the proposed expert-guided analysis enables a more meaningful spatially targeted parameter-space exploration. 
\autoref{tab:experts_cloverleaf} shows the per-expert PSNR and frequency measures averaged across all test members of the CloverLeaf3D dataset. Based on this frequency analysis, we rank the experts according to their focus on high- to low-frequency regions: $E3 > E2 > E1 > E4 \geq E5$.

\begin{figure}[htbp]
    \centering
    \noindent\includegraphics[width=1.0\linewidth]{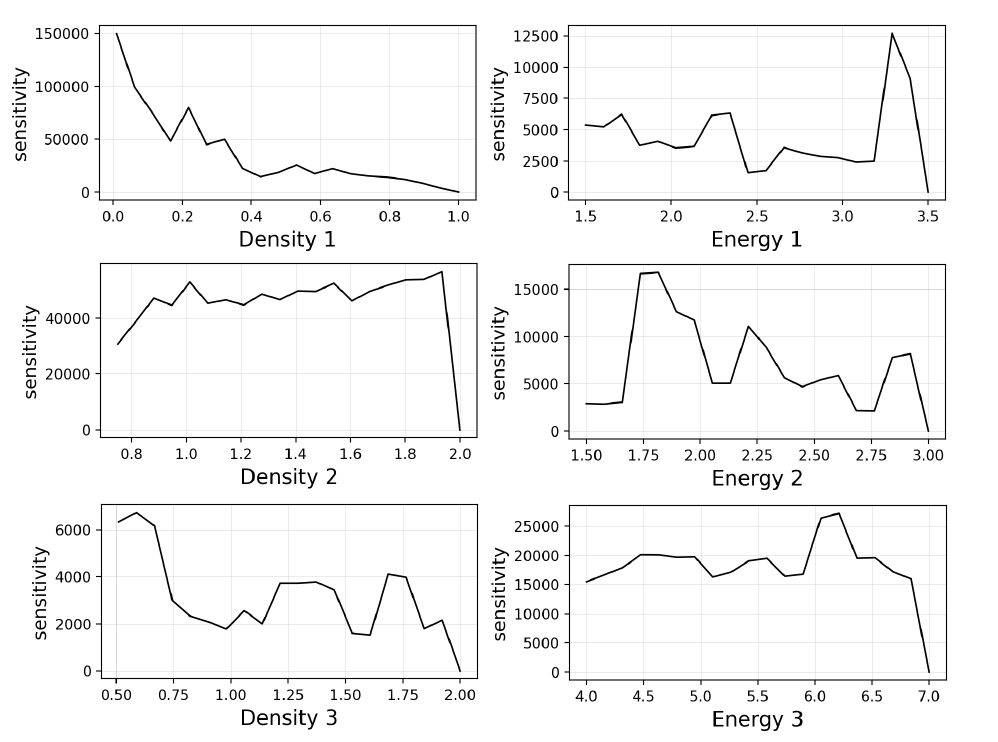}
    \caption{Parameter sensitivity of the six simulation parameters in the CloverLeaf3D dataset.}
    \label{fig:param_sensitivity_cloverleaf}
\end{figure}

Before investigating the expert-specific behaviors, we first analyze the global parameter sensitivity to understand which simulation parameters have the strongest overall influence on the CloverLeaf3D dataset. As shown in \autoref{fig:param_sensitivity_cloverleaf}, \textit{Density 1} and \textit{Density 2} present the largest responses among all six parameters, indicating that small fluctuations in these two density values lead to the most significant changes in the predicted hydrodynamic fields. Therefore, we focus on these two parameters for a more fine-grained expert-level analysis.
\par
Our goal is to determine whether this parameter sensitivity is spatially dependent. To this end, we plot the sensitivity curves of all experts under both \textit{Density 1} and \textit{Density 2} in \autoref{fig:P1P2_cloverleaf}. The results reveal a clear co-adaptive behavior between Expert-2 and Expert-5. Under both parameter sweeps, these two experts present highly similar response patterns, whereas the other experts follow entirely different trends.
This observation suggests that the spatial regions modeled by Expert-2 and Expert-5 may share similar physical responses to density-related changes, even though they correspond to different parts of the simulation domain.
More specifically, as \textit{Density 1} starts to increase, the field values associated with Expert-2 and Expert-5 increase sharply, while the regions captured by the other experts generally show the opposite trend. A similar pattern is also observed when sweeping Density 2, which further supports the hypothesis that these two experts are coupled in how they respond to changes in the parameter space.
Such region-specific behaviors would be difficult to identify through the conventional global sensitivity analysis, where the responses from the entire spatial domain are aggregated into a single summary curve.
\par
These observations highlight the value of our MoE-based design for scientific exploration. Because the learned experts provide a spatially specialized decomposition over the simulation domain, this allows scientists to move beyond global sensitivity summaries and inspect how different local regions respond to the same parameter.
These two case studies demonstrate that the learned expert assignment not only enhances modeling performance but also provides scientifically meaningful support for localized parameter-space exploration.

\begin{figure}[htbp]
    \centering
    \noindent\includegraphics[width=1.0\linewidth]{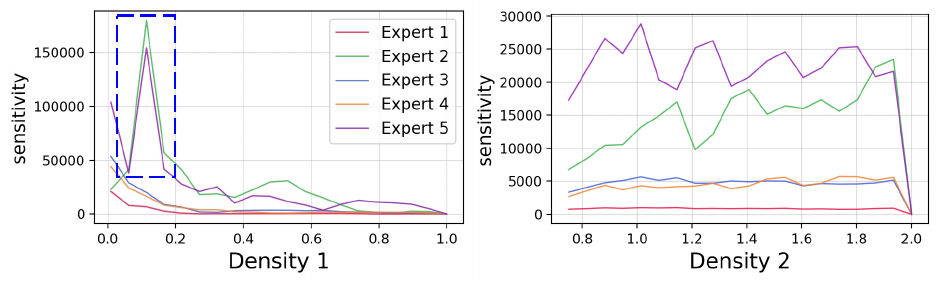}
    \caption{Per-expert sensitivity curves for the two most sensitive parameters in the CloverLeaf3D dataset. A clear co-adaptive pattern is observed between Expert-2 and Expert-5.}
    \label{fig:P1P2_cloverleaf}
\end{figure}


\section{Discussion and Future Work}
\label{sec:discussion}

In this section, we discuss the advantages of our proposed mixture-of-experts (MoE) architecture for INRs compared with the existing related frameworks, particularly in the context of scientific visualization. 
Existing MoE-based INR methods are designed solely based on MLPs, and they use MoE to improve the local function mapping capability~\cite{ben2024neural} or compression effectiveness~\cite{han2025moe}. Even though such a formulation is effective for general INR tasks, they still suffer from the limitations of fully ML-based INRs when modeling the scientific fields with localized high-frequency structures. 
\par
Moreover, our method does not require any form of pretraining for expert assignment. Existing MoE-based approaches adopt different strategies to stabilize the routing process. For example, Neural Experts~\cite{ben2024neural} is pretrained to begin with random uniform assignments, while MoE-INR~\cite{han2025moe} is pretrained with voxel clustering. Without an explicit pretraining strategy, their routing process can be unstable or even collapse in the early training stage. We attribute these behaviors to the limited spatial inductive bias of MLP-based gating networks. 
In contrast, our gating module employs a low-resolution feature grid that explicitly encodes the spatial information, which leads to a stable initialization and converges quickly in the early training stage. In addition, without relying on pretraining or pre-determining the expert assignment, our MoE framework can form a spatial partition that naturally reflects the true characteristics of the simulation domain. 
\par
Beyond the scalability offered by MoE architectures, our design also provides interpretable insights that support scientific exploration. By both visually and quantitatively associating each expert with frequency characteristics or specific scientific phenomena, we show that our learned expert assignment is scientifically meaningful, which has not been demonstrated in any existing MoE-based INR methods.
\par
Another line of INR work shares a similar motivation with our MoE design. To allocate more model capacity to the spatially complex regions, APMGSRN~\cite{wurster2023adaptively} achieves this by learning to transform multiple feature grids toward regions with high reconstruction errors. Each feature grid in their model plays a role similar to an expert in our framework. However, in APMGSRN, an entire feature grid can only scale, rotate, and translate as a whole, while our key-value pairs provide finer-grained adaptivity without wasting any representational capacity. Furthermore, grids that are initially placed in empty regions may receive little to no training signal, making their adaptive grids highly sensitive to initialization.
In contrast, our MoE-based architecture avoids this issue by learning a gating network to route coordinates to spatially specialized experts and leveraging cross-attention to further assign each coordinate to the corresponding feature vectors. 
\par
One limitation in our work is that our model requires slightly longer training time, particularly compared to the non-MoE INR models that rely on direct interpolation over the explicit structures, \eg, Explorable-INR and K-Planes. This overhead is primarily due to the use of cross-attention mechanisms and the MoE framework.
However, given the substantial time savings relative to running full numerical simulations, this additional training time is negligible in practice. For example, running 70 high-resolution MPAS-Ocean simulations takes approximately 83 hours, while training our model requires around 13 hours. When scaling to scenarios involving hundreds or thousands of simulations, this one-time training cost becomes minimal compared to the substantial time saved through surrogate predictions.

\section{Conclusion}

In this paper, we propose Feature-Adaptive INR (\ours), a compact embedding-augmented implicit neural representation for high-fidelity surrogate modeling of ensemble simulations. 
To address the rigidity of prior explicit feature representations, we introduce an adaptive encoding approach based on cross-attention with an augmented memory bank. 
To further effectively scale to the large complex ensemble datasets, we incorporate a Mixture-of-Experts (MoE) framework that dynamically routes inputs to relevant memory banks according to their spatial data characteristics. 
Beyond its technical contributions, our approach also enables interpretable predictions by showing which model components contribute to outputs at specific spatial locations. This can be achieved by first tracing the selected encoder experts and then the most highly activated key-value pairs. Such transparency is particularly valuable for scientific applications.
Extensive evaluations demonstrate that \ours achieves state-of-the-art surrogate modeling fidelity compared to existing INR-based surrogates. Moreover, through both quantitative metrics and visual analysis, we also showcase that our MoE design learns interpretable expert specialization that aligns with meaningful scientific structures.






\bibliographystyle{abbrv-doi-hyperref}

\bibliography{template}

\appendix 
\crefalias{section}{appendix} 







\end{document}